\documentclass[letterpaper]{article} 
\usepackage{aaai24}  
\usepackage{times}  
\usepackage{helvet}  
\usepackage{courier}  
\usepackage[hyphens]{url}  
\usepackage{graphicx} 
\usepackage{natbib}  
\usepackage{caption} 
\usepackage{multirow}
\usepackage{kotex}
\usepackage{booktabs}
\usepackage{amsmath}
\usepackage{amssymb}
\usepackage{comment}
\usepackage{pifont}
\usepackage{tabularx}
\newcolumntype{Y}{>{\centering\arraybackslash}X}

\usepackage{algorithm}
\usepackage{algorithmic}

\usepackage{newfloat}
\usepackage{listings}
\usepackage{xcolor}
\definecolor{codegreen}{rgb}{0,0.6,0}
\definecolor{codegray}{rgb}{0.5,0.5,0.5}
\definecolor{codepurple}{rgb}{0.58,0,0.82}

\lstdefinestyle{mystyle}{
    commentstyle=\color{codegreen},
    keywordstyle=\color{magenta},
    numberstyle=\tiny\color{codegray},
    stringstyle=\color{codepurple},
    basicstyle=\ttfamily\footnotesize,
    breakatwhitespace=false,         
    breaklines=true,                 
    captionpos=b,                    
    keepspaces=true,                 
    numbers=left,                    
    numbersep=5pt,                  
    showspaces=false,                
    showstringspaces=false,
    showtabs=false,                  
    tabsize=2
}

\DeclareCaptionStyle{ruled}{labelfont=normalfont,labelsep=colon,strut=off} 
\floatstyle{ruled}
\newfloat{listing}{tb}{lst}{}
\floatname{listing}{Algorithm}
\title{DiffRef3D: A Diffusion-based Proposal Refinement Framework for 3D Object Detection}
\author{
    Se-Ho Kim\equalcontrib,
    Inyong Koo\equalcontrib,
    Inyoung Lee,
    Byeongjun Park,
    Changick Kim
}
\affiliations{
    Korea Advanced Institute of Science and Technology\\

    \{ksh1040, iykoo010, inzero24, pbj3810, changick\}@kaist.ac.kr
}

\usepackage{bibentry}

\begin{document}

\maketitle

\begin{abstract}
Denoising diffusion models show remarkable performances in generative tasks, and their potential applications in perception tasks are gaining interest. 
In this paper, we introduce a novel framework named DiffRef3D which adopts the diffusion process on 3D object detection with point clouds for the first time. 
Specifically, we formulate the proposal refinement stage of two-stage 3D object detectors as a conditional diffusion process.
During training, DiffRef3D gradually adds noise to the residuals between proposals and target objects, then applies the noisy residuals to proposals to generate hypotheses.
The refinement module utilizes these hypotheses to denoise the noisy residuals and generate accurate box predictions. 
In the inference phase, DiffRef3D generates initial hypotheses by sampling noise from a Gaussian distribution as residuals and refines the hypotheses through iterative steps.
DiffRef3D is a versatile proposal refinement framework that consistently improves the performance of existing 3D object detection models.
We demonstrate the significance of DiffRef3D through extensive experiments on the KITTI benchmark.
Code will be available.
\end{abstract}

\section{Introduction}
Denoising diffusion models have demonstrated remarkable performance in the field of image generation \cite{ddpm_ho, ddim_song}.
Recently, the attempts to extend the use of the diffusion model to visual perception tasks \cite{dformer_wang, ddps_lai,diffusioninst_gu, d3dp_shan, diffusiondet_chen} have exhibited notable success. 
However, these successes are limited to perception tasks in image data, while the application to perception tasks involving other modalities (e.g., LiDAR point clouds) remains unsolved. 
In this paper, we introduce DIffRef3D, a novel framework that utilizes the diffusion process for the task of LiDAR-based 3D object detection.

\begin{figure}[t]
    \centering
    \includegraphics[width=\columnwidth]{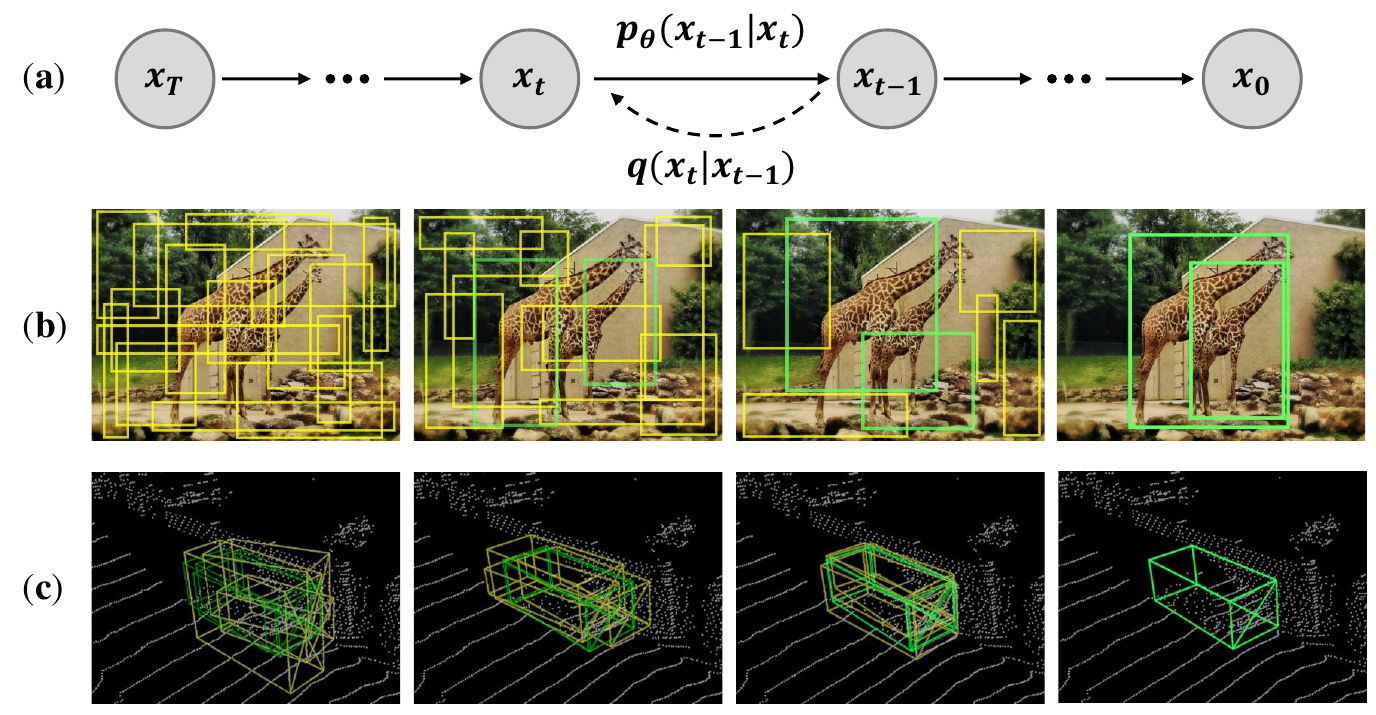}
    \caption{\textbf{Diffusion model for 2D and 3D object detection}: (\textbf{a}) A diffusion model where $q$ and $p_\theta$ are the forward and reverse diffusion process respectively. (\textbf{b}) Diffusion model for 2D object detection (DiffusionDet). (\textbf{c}) We formulate 3D object detection as a conditional diffusion process based on the region proposal. Noisy boxes (or hypotheses) are colored in yellow, and predictions are colored in green.}
    \label{fig1_overview}
\end{figure}

DiffRef3D is mostly inspired by DiffusionDet \cite{diffusiondet_chen} which aims to solve 2D object detection by adopting the diffusion process to Sparse R-CNN \cite{sparsercnn_sun}. 
The key concept of DiffusionDet is to scatter random noisy boxes along the image scene and gradually refine the noisy boxes to the object bounding boxes by the reverse diffusion process.
However, a straightforward extension of the DiffusionDet pipeline to the 3D domain encounters challenges attributed to the inherent dissimilarity between image and point cloud data.
Firstly, since an image is consecutive grid-structured data, meaningful features can be extracted from a random location. 
In contrast, the random placement of boxes in 3D scenes often falls short of providing any information due to the sparsity of point cloud data caused by occlusion and the inherent limitation of LiDAR sensors. 
Secondly, while 2D objects have various sizes in an image frame due to perspective, objects in point cloud data are relatively small compared to the 3D scene.
Given these characteristics of 3D scenes, scattering noisy boxes for object detection without any prior is akin to finding a needle in a haystack.
Inevitably, the diffusion model on 3D point cloud space requires extended prerequisites such as coarse information about object sizes and displacements.
Therefore, we design DiffRef3D to be a proposal refinement framework that enhances two-stage detectors, leveraging guidance of the region proposals generated from the first stage.

In DiffRef3D, we formulate the proposal refinement stage of two-stage detection models as a conditional diffusion process. 
For each proposal, our framework applies a noisy box residual and generates a hypothesis. 
The model is trained to denoise the residuals by the reverse diffusion process and predict object bounding boxes, using the knowledge of both the proposal and hypothesis regions.
Figure~\ref{fig1_overview} highlights conceptual differences between DiffusionDet and DiffRef3D, which are proposed for 2D and 3D object detection tasks, respectively.
Specifically, DiffRef3D employs region proposals as the condition for both its forward and reverse diffusion process, as opposed to DiffusionDet, which is built upon an unconditional diffusion process.
To this end, we introduce a hypothesis attention module (HAM), which treats features extracted from hypotheses to be conditioned on region proposals.
During training, the noisy residuals are generated through the forward diffusion process from the true residuals between the proposals and target object bounding boxes. 
For inference, the residuals are sampled from a Gaussian distribution, and the model iteratively refines the prediction by denoising the hypotheses. 
We adopt the iterative sampling method from DDIM \cite{ddim_song} to predict intermediate hypotheses during inference. 
DiffRef3D can be implemented with existing two-stage 3D object detectors and consistently improve performances compared to the baseline.
We demonstrate the performance and scalability of DiffRef3D through extensive experiments held on the KITTI benchmark \cite{kitti}.

In summary, our main contributions are as follows:
\begin{itemize}
    \item DiffRef3D serves as a generalizable proposal refinement framework that can be applied to existing two-stage 3D object detectors, resulting in consistent performance improvements.
    \item To the best of our knowledge, DiffRef3D represents the pioneering effort in employing the diffusion process for 3D object detection, interpreting the proposal refinement stage as a conditional diffusion process.
\end{itemize}

\section{Related Works}
\subsection{LiDAR-based 3D Detection}
3D object detectors can be broadly categorized into two methods: point-based methods and voxel-based methods.
Point-based methods \cite{pointnet_qi, pointnet++_qi} take the raw point clouds as inputs and employ permutation invariant operations to aggregate the point features.
Voxel-based methods \cite{voxelnet_zhou, second_yan} divide point clouds into a regularly structured grid and apply convolutional operations to extract features.
Two-stage models employ the above methods to serve as a region proposal network (RPN), introducing the proposal refinement stage that further improves the prediction quality.

PointRCNN \cite{pointrcnn_shi} performs foreground segmentation and generates initial proposals using a PointNet++ \cite{pointnet++_qi} backbone, and then extracts RoI features from the point clouds inside the proposals for refinement. 
PV-RCNN \cite{pvrcnn_shi} proposes a Voxel Set Abstraction module to encode the voxel-wise feature volumes into a small set of keypoints, then aggregates to the RoI-grid points to take advantage of both point-based and voxel-based methods to refine proposals. 
Voxel R-CNN \cite{voxelrcnn_deng} follows a typical pipeline of anchor-based two-stage models, using SECOND \cite{second_yan} as RPN and refine proposals with the voxel features pooled for RoI.
CenterPoint \cite{centerpoint_yin} predicts proposals from object centers computed from class-specific heatmap then point features are extracted from 3D centers of each face of the proposals for refinement.
CT3D \cite{ct3d_sheng} embeds and decodes proposals into effective proposal features based on neighboring point features for box prediction by channel-wise transformer.

There are special cases of two-stage models which utilize additional proposal refinement modules in order to adopt the cascade detection paradigm.
3D Cascade RCNN \cite{3dcascadercnn_cai} iteratively refines proposals through cascade detection heads while considering point completeness score.
CasA \cite{casa_wu} progressively refines proposals by cascade refinement network while aggregating features from each stage by cascade attention module.
The previous cascade detection paradigm boosts performance by leveraging the advantages of multiple predictions such as ensemble, however it lacks flexibility in sampling steps and increases memory usage for additional detection heads.
In contrast, iterative sampling of diffusion-based models exploits advantages without increasing memory usage for each sampling step and sampling steps are adjustable without additional training.

\subsection{Diffusion Framework on Perception Task}
The diffusion framework \cite{adm_dhariwal, ddpm_ho} has shown remarkable performance in the field of image generation. 
Notably, DDIM \cite{ddim_song} expedites the sampling process through the utilization of a more efficient class of iterative implicit probabilistic models.
Drawing attention to its denoising capabilities and performance, there have been exploding recent efforts to apply it to perception tasks \cite{d3dp_shan, diffbev_zou} beyond generation tasks. 
DiffusionDet \cite{diffusiondet_chen} first adopt the diffusion model to the 2D object detection problem by denoising random boxes to object bounding boxes via the diffusion process. 
DiffusionInst \cite{diffusioninst_gu} utilizes additional mask branches and instance-aware filters upon DiffusionDet to address the instance segmentation task, and the following works \cite{ddps_lai, dformer_wang} show competitive performance on semantic segmentation. 
However, diffusion-based perception models have been mostly explored in the image domain.  
Through DiffRef3D, we demonstrate the first application of the diffusion framework to a 3D perception task that involves point cloud input. 
\section{Preliminaries}

\subsection{Proposal Refinement Module}
A proposal refinement module in two-stage 3D object detectors takes the initial proposals of the RPN as input and exploits the information of the RoIs to refine them.
Formally, the refinement module takes a proposal $X^P$ and outputs the box residual $\hat{x}$ and the confidence score $\hat{c}$:
\begin{align}
    f^P &= \textrm{RoIPool}(X^P), \label{eq:roipool}\\
    \hat{x}, \hat{c} &= \textrm{Det}(f^P). \label{eq:detection_head}
\end{align}

The RoI feature pooling operation $\textrm{RoIPool}(\cdot)$ extracts proposal feature $f^P$ from $X^P$.
$\textrm{Det}(\cdot)$ represents the detection head, which consists of a regression branch producing $\hat{x}$ and a classification branch producing $\hat{c}$.
$\hat{x}$ is supervised with the target residual $x^{gt}$ between the proposal and its target object box $X^T$, i.e., $x^{gt} = X^T \ominus X^P$ where $\ominus$ represents box encoding functions \cite{second_yan}, and the target for $\hat{c}$ is calculated with the intersection over union (IoU) between $X^P$ and $X^{T}$. 

\subsection{Conditional Diffusion Model}
During training, a small amount of Gaussian noise is gradually added to the data sample $x_0$ through a Markovian chain of the forward diffusion process. 
The forward diffusion process is formulated as:
\begin{equation}
\label{eq:forward_one_step}
    q(x_t|x_{t-1})=\mathcal{N}(x_t;\sqrt{1-\beta_t}x_{t-1},\beta_t \mathbf{I}),
\end{equation}
where $\beta_t$ represents a noise variance schedule \cite{ddpm_ho, improved_nichol}. 
For brevity, let $\alpha_t=1-\beta_t$, $\bar{\alpha_t}=\prod_{i=1}^t \alpha_i$, and a noisy sample $x_t$ at timestep $t, 0\leq t \leq T,$ is derived as follows:
\begin{equation}
\label{eq:forward_t_step}
    q(x_t|x_0)=\mathcal{N}(x_t;\sqrt{\bar{\alpha_t}},(1-\bar{\alpha_t})\mathbf{I}),
\end{equation}
\begin{equation}
\label{eq:data_to_noise}
    x_t=\sqrt{\bar{\alpha_t}}x_0+(1-\bar{\alpha_t})\epsilon, \textnormal{where } \epsilon\sim\mathcal{N}(0, \mathbf{I}).
\end{equation}

The conditional diffusion model learns a reverse diffusion process, aiming to progressively refine $x_t$ back to $x_0$ with the guidance of the conditional input $y$. 
The reverse diffusion process is formulated as follows:
\begin{equation}
\label{eq:reverse_process}
    p_\theta(x_{0:T}|y)=p(x_T)\prod^T_{t=1}p_\theta(x_{t-1}|x_t,y).
\end{equation}

\section{DiffRef3D}
In this section, we introduce DiffRef3D, a diffusion-based proposal refinement framework for 3D object detection.
We formulate the refinement module of two-stage detectors as a conditional diffusion model that takes proposal $X^P$ as the conditional input and denoises the noisy residual $x_t$ sampled from timestep $t$.
DiffRef3D reconstructs the actual residual towards the object $\hat{x}_0^{(t)}$ by the reverse diffusion process and also outputs the confidence score $\hat{c}^{(t)}$:
\begin{equation}
\label{eq:diffref3d_module}
    \hat{x}_0^{(t)}, \hat{c}^{(t)} = \textrm{DiffRef3D}(x_t, t |X^P).
\end{equation}

\begin{figure}
    \centering
    \includegraphics[width=\columnwidth]{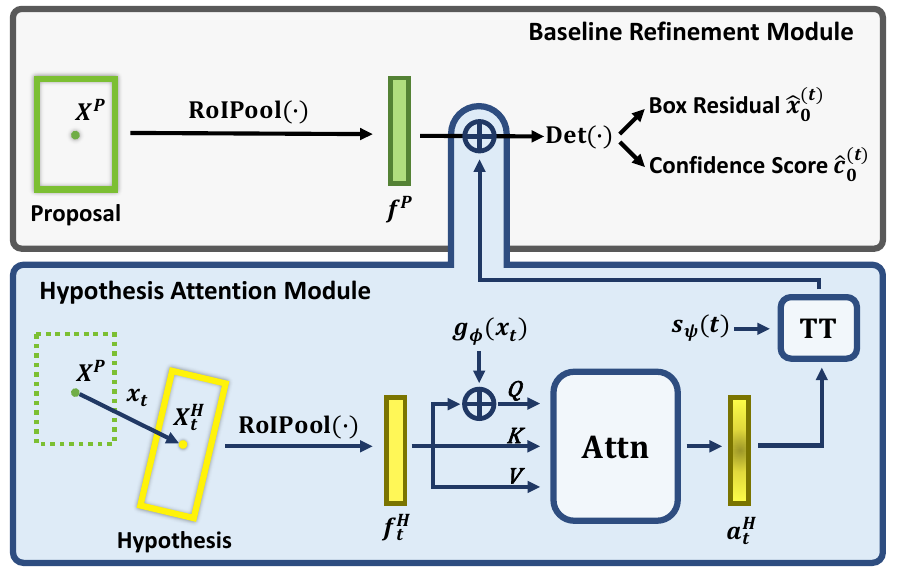}
    \caption{\textbf{Diffusion-based Refinement Module.} Based on an existing proposal refinement module, DiffRef3D further explores a secondary region (hypothesis) feature by processing through the hypothesis attention module.}
    \label{fig2_architecture}
\end{figure}

As shown in Fig. \ref{fig2_architecture}, DiffRef3D can be implemented on any existing refinement module that follows the general formulation of Eq. \ref{eq:roipool}, \ref{eq:detection_head} by adding a simple parallel module named the hypothesis attention module (HAM).
Details of the HAM architecture, training, and inference procedures are described in the following subsections. 

\subsection{Hypothesis Attention Module}
The hypothesis $X^H_t$ is a secondary inspection region derived by applying the given box residual to the proposal. i.e., $X^H_t = X^P \oplus x_t$.
As the baseline refinement module extracts the proposal feature $f^P$ (Eq. \ref{eq:roipool}), HAM also extracts the hypothesis feature $f^H_t$ using the same RoI feature pooling operation:
\begin{equation}
\label{eq:hypothesis_pooling}
    f^H_t = \textrm{RoIPool}(X^H_t).
\end{equation}
The hypothesis feature is further processed to incorporate the information of $x_t$ and $t$ by applying a self-attention block $\textrm{Attn}(Q, K, V)$ and a temporal transformation block $\textrm{TT}(\cdot, t)$.

The self-attention block is a common multi-head attention module \cite{transformer_vaswani} that takes $f^H_t$ as input for the query, key, and value. 
Except, we embed $x_t$ as a positional encoding to the query. 
We utilize a shallow MLP $g_\phi(\cdot)$ to convert $x_t$ as the same dimensional vector with $f^H_t$.
Formally, the attention feature of the hypothesis $a^H_t$ is computed as:
\begin{equation}
\label{eq:self_attention_block}
    a^H_t = \textrm{Attn}(f^H_t+g_\phi(x_t), f^H_t, f^H_t).
\end{equation}
$a^H_t$ is a feature vector that represents the information of the hypothesis region with consideration of its relative displacement from the proposal. 

The temporal transformation block reweights $a^H_t$ with respect to $t$ to adjust the effect of the noise signal which gets dominant as timestep increases.
Specifically, the output of HAM $h_t$ is formulated as follows:
\begin{equation}
\label{eq:temporal_transform_block}
    h_t = \textrm{TT}(a^H_t, t) = W_t \cdot a^H_t + b_t
\end{equation}
where $W_t, b_t$ are the scaling and shifting factors derived from an MLP $s_\psi(\cdot)$:
\begin{equation}
\label{eq:scale_shift}
    [W_t; b_t] = s_\psi(t).
\end{equation}

Finally, $h_t$ is aggregated with the proposal feature for prediction:
\begin{equation}
\label{eq:diffusion_detection_head}
    \hat{x}^{(t)}_0, \hat{c}^{(t)} = \textrm{Det}(f^P + h_t).
\end{equation}

\subsection{Training}
During the training stage, We set the reconstruction target $x_0=x^{gt}$ and sample noisy residual $x_t$ by the forward diffusion process. 
The model is trained to perform the reverse diffusion process of reconstructing the true residual, which is identical to refining a hypothesis to the target object bounding box.
We elucidate the methodology for generating hypotheses from proposals and present the overall training process of DiffRef3D in pseudo-code through Algorithm \ref{lst:train_procedure}.

\begin{listing}[tb]
\caption{DiffRef3D training procedure}
\label{lst:train_procedure}
\begin{lstlisting}[language=Python]
def train(pr_box, gt_box):
  # pr_box: proposal box [B, N, 7]
  # gt_box: ground truth box [B, *, 7]

  # Compute residual and class label
  x_0, c_gt = \
    assign_target(pr_box, gt_box)
    
  t = randint(0,T)
  eps = normal(mean=0,std=1) # [B, N, 7]
  SNR = 2
  x_t = sqrt(1 - alpha_bar_t) * x_0 +
        (1 - alpha_bar_t) * eps / SNR
        
  # Predict
  x_hat, c_hat = \
    DiffRef3D(pr_box, x_t, t)

  # Compute task specific losses
  reg_loss = get_reg_loss(x_0, x_hat)
  cls_loss = get_cls_loss(c_gt, c_hat)
  return reg_loss + cls_loss
\end{lstlisting}
\end{listing}

\subsubsection{Hypothesis generation}
From our attempts, spreading random boxes along the point cloud scene to detect objects fails to train 3D object detection by the diffusion process.
Therefore, we propose hypothesis generation to map random signals into 3D boxes rather than mapping normalized values directly into a point cloud scene.
Utilizing proposals as a reference frame to generate a hypothesis guarantees that random boxes start in close proximity to the potential object locations.
Before we apply the forward diffusion process on the target residual $x^{gt}$, normalization is preceded since the range of the residual differs according to the size of each proposal.
Details of the normalization process are provided in supplementary materials.
Noise scaled by signal-to-noise ratio (SNR) is applied to $x_0$ according to Eq. \ref{eq:data_to_noise} results in noisy residual $x_t$.
Finally, hypothesis $X^H_t$ is obtained by applying noisy residual $x_t$ on proposal $X^P$.

\subsubsection{Training objectives}
Recent diffusion models \cite{ddpm_ho} are trained with L2 loss between predicted noise and actual noise added to the data sample. 
However, noise prediction could be replaced by direct prediction of data samples. 
A number of diffusion-based approaches \cite{diffusiondet_chen, diffusioninst_gu, ddp_ji} on perception tasks report that it is better to predict data samples and train on task-specific losses. 
Following consensus from previous works, we also trained our model with task-specific losses, e.g., binary cross-entropy loss for classification, and smooth-L1 loss for regression.

\subsection{Inference}
During the inference stage, DiffRef3D refines the hypothesis generated from random Gaussian noise and follows iterative sampling steps from DDIM \cite{ddim_song} for progressive refinement.
Starting from a random Gaussian noise $x_T\sim\mathcal{N}(0,1)$, DDIM progressively denoises Gaussian noise into data samples with step size $\Delta t$, i.e., $x_T\rightarrow x_{T-\Delta t} \rightarrow ... \rightarrow x_0$.
The progressive refinement is formulated as follows:
\begin{equation}
    \label{eq:iterative_sampling}
    x_{t-\Delta t} = \hat{x}^{(t)}_0\sqrt{\alpha_{t-\Delta t}} + \epsilon^{(t)}\sqrt{1-\alpha_{t-\Delta t}-\sigma^2_t}+\sigma_t\epsilon
\end{equation}
\begin{align}
    \label{eq:noise_prediction}
\epsilon^{(t)}&=\frac{x_t-\alpha_t\hat{x}^{(t)}_0}{\sqrt{1-\alpha_t}} \\
\sigma_t&=\sqrt{(\frac{1-\alpha_{t-\Delta t}}{1-\alpha_t})}\sqrt{(1-\frac{\alpha_t}{\alpha_{t-\Delta t}})}
\end{align}
where $\epsilon^{(t)}$ represents noise term added to data sample and $\sigma_t$ represents stochastic parameter, respectively. Algorithm \ref{lst:inference_procedure} describes the inference procedure of DiffRef3D in pseudo-code.

\begin{listing}
\caption{DiffRef3D inference procedure}
\label{lst:inference_procedure}
\begin{lstlisting}[language=Python]
def inference(pr_box, steps):
  # pr_box: proposal box [B, N, 7]
  # steps: sampling steps
  T = 1000

  t = reversed(linspace(-1,T,steps))
  ts = list(zip(times[:-1], times[0:]))
  
  x_t = normal(mean=0,std=1) # [B, N, 7]

  for t_now, t_next in zip(ts):
    # Predict at timestep t_now
    x_hat, c_hat = \
      DiffRef3D(pr_box, x_t, t_now)

    pr_box = \
      apply_residual(pr_box, x_hat)

    # Estimate x_t at timestep t_next
    x_t = \
    ddim_step(x_t, x_hat, t_now, t_next)
  return x_hat, c_hat
\end{lstlisting}
\end{listing}

\subsubsection{Iterative sampling}

\begin{table*}[ht!]
 \caption{Comparison with baseline models on the KITTI \textit{val} set}
  \centering
  \label{table:kitti_val}
  \begin{tabularx}{0.9\textwidth}{c |YYY|YYY|YYY}
    \toprule
    \multirow{2}{*}{Method} & 
    \multicolumn{3}{c|}{Car 3D AP$_{R40}$} &
    \multicolumn{3}{c|}{Ped. 3D AP$_{R40}$} &
    \multicolumn{3}{c}{Cyc. 3D AP$_{R40}$} \\
     & Easy & Mod. & Hard & Easy & Mod. & Hard & Easy & Mod. & Hard  \\
     \midrule
     Voxel R-CNN \cite{voxelrcnn_deng} & 92.24 & 84.57 & 82.30 & 64.97 & 58.07 & 52.66 & 91.83 & 73.13 & 68.40 \\ 
     DiffRef3D-V (1 step)& 92.53 & 84.98 & 82.55 & 67.32 & 60.45 & 55.28 & 89.13 & 71.80 & 67.60 \\
     DiffRef3D-V (3 step)& 92.89 & 84.77 & 82.64 & 69.70 & 63.07 & 57.70 & 93.33 & 74.20 & 69.63 \\
     \midrule
     PV-RCNN \cite{pvrcnn_shi}& 92.10 & 84.36 & 82.48 & 64.26 &56.67&51.91&88.88&71.95&66.78 \\ 
     DiffRef3D-PV (1 step)& 91.67 & 84.21 & 81.93 & 64.56 & 58.79 & 54.44 & 92.37 & 72.44 & 68.42  \\
     DiffRef3D-PV (3 step)& 91.84 & 84.33 & 82.15 & 65.71 & 59.46 & 55.13 & 93.76 & 73.94 & 69.60 \\
     \midrule
     CT3D \cite{ct3d_sheng} & 92.34&84.97& 82.91 &	61.05&55.57&51.10&89.01&71.88&67.91 \\
     DiffRef3D-T (1 step)& 92.99	& 85.13	& 82.83	& 65.09 & 59.13 & 54.36 & 88.20 & 71.27 & 67.13 \\
     DiffRef3D-T (3 step)& 93.28 & 85.13 & 82.97 & 67.49 & 61.29 & 56.17 & 90.92 & 72.61 & 68.52 \\
    \bottomrule
  \end{tabularx}
\end{table*}

DiffRef3D generates hypotheses in the first sampling step using residuals sampled from a Gaussian distribution. 
These hypotheses perform bounding box predictions in conjunction with proposal features, and the predicted outcomes serve two purposes: firstly, as the subsequent proposal in the next sampling step, and secondly, they contribute to inferring the next-step hypothesis according to Eq. \ref{eq:iterative_sampling}.
The iterative sampling steps share similarities with the cascade detection paradigm \cite{3dcascadercnn_cai, casa_wu}; however, in the case of cascade detection, distinct weights for each iteration's detection head are required, limiting the number of iterations. 
In contrast, iterative sampling employs a single-weight detection head, minimizing memory consumption and allowing customizable sampling repetitions.

\section{Experiments}
In this section, we delve into the intricate implementation details and outline the evaluation setup adopted to assess the performance of DiffRef3D.
Moreover, we conduct ablation studies to comprehensively examine each element of DiffRef3D and affirm the validity of our design choices.

\subsection{Implementation Details}
\subsubsection{Architecture}
We implement DiffRef3D on three popular two-stage 3D object detectors: Voxel R-CNN \cite{voxelrcnn_deng}, PV-RCNN \cite{pvrcnn_shi}, and CT3D \cite{ct3d_sheng}.
The models enhanced with DiffRef3D framework are denoted as DiffRef3D-V, DiffRef3D-PV, and DiffRef3D-T, respectively.
The model architectures mostly follow the original baseline, and only the HAM module is newly introduced. Given the $\textrm{RoIPool}(\cdot)$ output feature dimension $d$, ($d=96$ for Voxel R-CNN, 128 for PV-RCNN and CT3D), $a^H_t$ and $h_t$ are also $d$-dimensional vectors. $\textrm{Attn}(\cdot)$ is a multi-head attention module with the head number of 8, and $g_\phi(\cdot)$ is a two-layer MLP with the hidden dimension of $d$. $s_\psi(\cdot)$ is also a two-layer MLP, with the hidden dimension of $4d$.

\subsubsection{Training}
For consistency, training configurations for the 3D object detectors mirror those reported in the respective baseline works. 
In the context of the diffusion process, we set the maximum timestep $T$ to 1000 and we adopt a cosine noise schedule \cite{improved_nichol} for the forward diffusion process.
From our observations, SNR for signal scaling results in optimal performance at 2.
During the training stage, only one hypothesis was generated for each proposal.
All models were trained using 4 NVIDIA RTX 3090 GPUs, and the training duration was largely consistent with the baseline models due to the limited computational overhead introduced by the HAM.

\subsubsection{Inference}
In the inference phase, DiffRef3D can control the number of sampling steps and the number of hypothesis generation for each proposal.
However, we observed that increasing the number of hypotheses yields only minimal performance improvement. 
Therefore, we fixed the number of hypothesis generation for each proposal as one.
We also employ an ensemble method to make usage of the output of intermediate steps by averaging the predictions.

\subsection{Results on KITTI Benchmark}

\begin{table*}[ht!]
 \caption{Comparison with state-of-the-art methods on the KITTI \textit{test} set}
  \centering
  \label{table:kitti_test}
  \begin{tabularx}{0.9\textwidth}{c |Y Y Y|Y Y Y|Y Y Y}
    \toprule
    \multirow{2}{*}{Method} & 
    \multicolumn{3}{c|}{Car 3D AP$_{R40}$} &
    \multicolumn{3}{c|}{Ped. 3D AP$_{R40}$} &
    \multicolumn{3}{c}{Cyc. 3D AP$_{R40}$} \\
     & Easy & Mod. & Hard & Easy & Mod. & Hard & Easy & Mod. & Hard  \\
     \midrule
     SECOND \cite{second_yan} & 83.34 & 72.55 & 65.82 & 48.73 & 40.57 & 37.77 & 71.33 & 52.08 & 45.83 \\
    PointPillars \cite{pointpillars_lang}& 82.58 & 74.31 & 68.99 & 51.45 & 41.92 & 38.89 & 77.10 & 58.65 & 51.92 \\
    PointRCNN \cite{pointrcnn_shi}& 86.96 & 76.50 & 71.39 & 47.98 & 39.37 & 36.01 & 74.96 & 58.82 & 52.53 \\
    STD \cite{std_yang}& 86.61 & 77.63 & 76.06 & 53.08 & 44.24 & 41.97 & 78.89 & 62.53 & 55.87 \\
    Point-GNN \cite{pointgnn_shi}& 88.33 & 79.47 & 72.29 & 51.92 & 43.77 & 40.14 & 78.60 & 63.48 & 57.08 \\
    3DSSD \cite{3dssd_yang}& 88.36 & 79.57 & 74.55 & 50.64 & 43.09 & 39.65 & 82.48 & 64.10 & 56.90 \\
    SA-SSD \cite{sassd_he}& 88.75 & 79.79 & 74.16 & $-$ & $-$ & $-$ & $-$ & $-$ & $-$ \\
    PV-RCNN \cite{pvrcnn_shi}& 90.25 & 81.43 & 76.82 & 52.17 & 43.29 & 40.29 & 78.60 & 63.71 & 57.65 \\
    Part-A2 \cite{parta2_shi}& 87.81 & 78.49 & 73.51 & 53.10 & 43.35 & 40.06 & 79.17 & 63.52 & 56.93 \\
    Voxel R-CNN \cite{voxelrcnn_deng} & 90.90 & 81.62 & 77.06 & $-$ & $-$ & $-$ & $-$ & $-$ & $-$ \\
    VoTr \cite{votr_mao}& 89.90 & 82.09 & 79.14 & $-$ & $-$ & $-$ & $-$ & $-$ & $-$ \\
    SPG \cite{spg_xu}& 90.64 & 82.66 & 77.91 & $-$ & $-$ & $-$ & 80.21 & 66.96 & 63.61 \\
    CT3D \cite{ct3d_sheng} & 87.83 & 81.77 & 77.16 & $-$ & $-$ & $-$ & $-$ & $-$ & $-$ \\
    BTCDet \cite{btc_xu}& 90.64 & 82.86 & 78.09 & $-$ & $-$ & $-$ & 82.81 & 68.68 & 61.81 \\
    PDV \cite{pdv_hu}& 90.43 & 81.86 & 77.36 & 47.80 & 40.56 & 38.46 & 83.04 & 67.81 & 60.46 \\
    3D Cascade RCNN \cite{3dcascadercnn_cai} & 90.46 & 82.16 & 77.06 & $-$ & $-$ & $-$ & $-$ & $-$ & $-$ \\
    
    PG-RCNN \cite{pgrcnn_koo} & 89.38 & 82.13 & 77.33 & 47.99 & 41.04 & 38.71 & 82.77 & 67.82 & 61.25 \\
    \midrule
    DiffRef3D-V & 90.45& 81.29& 76.66& 46.59& 40.55& 38.27&80.16&66.61&59.98\\
    \bottomrule
\end{tabularx}
\end{table*}
The KITTI benchmark \cite{kitti} is one of the most popular benchmarks for 3D object detection evaluation.
The dataset consists of 7,481 LiDAR samples for training and 7,518 LiDAR samples for testing. 
Training samples are divided into the \textit{train} set with 3,712 samples and the \textit{val} set with the remaining 3,769 samples.
We present two performance evaluations for comparative analysis: For evaluation on KITTI \textit{val} set, we compared DiffRef3D-V, DiffRef3D-PV, and DiffRef3D-T trained with the \textit{train} set with their respective baselines. For KITTI \textit{test} set, we trained a model with our framework using randomly selected 5,981 samples, validated with the remaining 1,500 samples, and evaluated the performance on the online test leaderboard.  
The performance were assessed across three classes: \textit{car}, \textit{pedestrian}, and \textit{cyclist} on three different levels (easy, moderate, and hard).
The results on both the \textit{val} and \textit{test} sets were evaluated using the mean average precision calculated at 40 recall positions.

\subsubsection{Comparison with baseline models}
Table \ref{table:kitti_val} outlines the efficacy of DiffRef3D implementation across three baseline models on the KITTI \textit{val} set.
We present the performance of the baseline model by incorporating the officially released model weights from each study except Voxel R-CNN which is reported without \textit{pedestrian} and \textit{cyclist} classes results.
Therefore we reproduced Voxel R-CNN detection results on three classes based on their open-source code.
DiffRef3D improves detection performance for \textit{pedestrian} and \textit{cyclist} classes for all baseline models and difficulties, while performance remains consistent or decreases slightly in \textit{car} class.
We consider this is due to \textit{car} class proposal boxes being sufficiently large to exploit local features which results DiffRef3D performs similar to baseline models on \textit{car} class detection.
DiffRef3D with sampling steps of 3 improves the baseline Voxel R-CNN, PV-RCNN, and CT3D with 5.00\%, 2.79\%, and 5.72\% AP (R40), respectively, in the moderate \textit{pedestrian} class.
Moreover, DiffRef3D improves the \textit{cyclist} detection performance by 1.07\%, 1.99\%, and 0.73\%, respectively.
The performance improvements primarily stem from the hypothesis generation, which contributes additional local features from individual proposals, consequently yielding greater enhancements in the \textit{pedestrian} and \textit{cyclist} classes.
Notably, as the sampling step increases in DiffRef3D, detection performance increases in the trade-off of computation costs.
Nevertheless, the performance improvement becomes less pronounced relative to the escalating computational demands as the sampling step is raised. 
Therefore, we present the results of the sampling steps of 1 and 3 in comparison with baseline models.

\subsubsection{Comparison with state-of-the-art methods}
To further comparisons with state-of-the-art methods, we report results of DiffRef3D-V on the KITTI \textit{test} set since it shows the best overall performances among three baseline enhanced models.
Therefore we trained DiffRef3D-V on a random split of training samples and reported results on the KITTI \textit{test} set with a sampling step of 3.
Table \ref{table:kitti_test} summarizes the comparison with state-of-the-art methods.
DiffRef3D-V shows comparable performance in \textit{car} and \textit{cyclist} with other models, while performance lags slightly on \textit{pedestrian}.
Nevertheless, as DiffRef3D is applicable to other two-stage 3D detectors published later, it can consistently show competitive performance with state-of-the-art models.

\subsection{Ablation study}
We conduct extensive ablation studies on the KITTI \textit{val} set to verify our design choices of DiffRef3D.

\begin{figure*}[t]
    \centering
    \includegraphics[width=\textwidth]{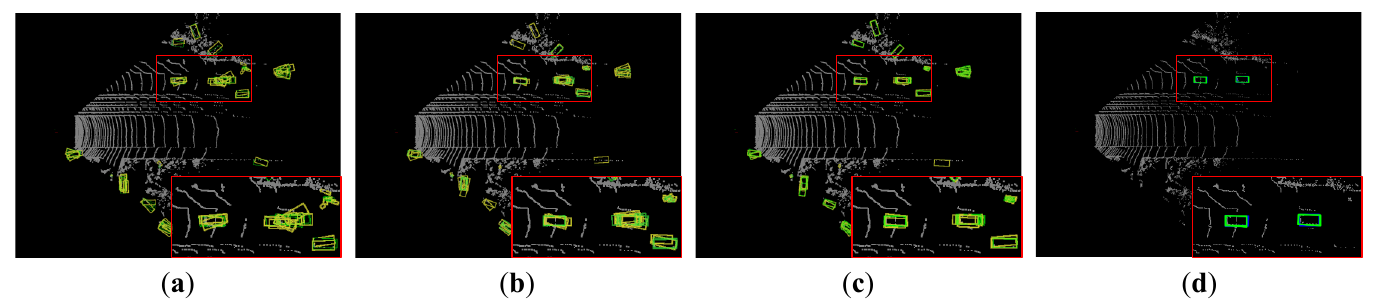}
    \caption{\textbf{Visualization of iterative proposal refinement on the KITTI benchmark}. 
    DiffRef3D refines the proposals using features from the previous predictions and the hypotheses sampled at the current timestep. Each image shows hypotheses and predictions at (a) t=1000, (b) t=500, (c) t=200, and (d) the final prediction. Hypotheses are colored in yellow, predictions are colored in green, and ground truths are colored in blue.}
    \label{fig3_kitti_visualization}
\end{figure*}

\subsubsection{Effectiveness of HAM and diffusion process}
\begin{table}[ht!]
    \centering
    \begin{tabularx}{0.9\columnwidth}{YY|YYY}
    \toprule
    \multirow{2}{*}{HAM} &
    \multirow{2}{*}{D.P.} & 
    \multicolumn{3}{c}{Sampling Steps} \\
    & & 1 & 2 & 3 \\
    \midrule
              &  & 59.62 & 61.01 & 61.38 \\
    \ding{51} &  & 59.22 & 60.91 & 61.50\\
    \ding{51} & \ding{51} & 60.45 & 62.77 & 63.07 \\
    \bottomrule
    \end{tabularx}
    \caption{Ablation studies on the KITTI \textit{val} set with AP (R40) for pedestrian in moderate difficulty with different configurations .}
    \label{table:ablation_ham_dp}
\end{table}

To investigate the effects of HAM and the diffusion process (denoted as D.P. in Table \ref{table:ablation_ham_dp}), we assessed performance by implementing HAM and D.P. with Voxel R-CNN as our baseline model.
DiffRef3D-V refers to configuration with HAM and D.P. all enabled.
During ablation study, box ensemble is applied as default in multiple sampling steps.
Moreover, we also applied iterative update of box predictions on baseline either for fair comparison.
Table \ref{table:ablation_ham_dp} summarizes the result of the experiment.
In the absence of the D.P., which implies that hypotheses are sampled from random Gaussian noise during both training and testing, AP (R40) decreased by 0.4\% and 0.1\% in sampling steps of 1 and 2, respectively.
This suggests hypotheses generated by noise that does not follow the diffusion process distracts baseline refinement module.
Integrating all of our proposed components results in AP (R40) improvements compared to baseline with 0.83\%, 1.76\%, and 2.69\% for each sampling steps, respectively.
Note that performance improvement introduced by iterative sampling steps is higher with D.P., baseline AP (R40) increased by 0.39\% and DiffRef3D-V increased by 2.32\% at sampling steps of 2 compared to sampling steps of 1.

\subsubsection{Sampling steps}
\begin{table}[ht!]
    \centering
    \begin{tabularx}{0.9\columnwidth}{c|YYY|c}
    \toprule
    \multirow{2}{*}{Sampling Steps} &
    \multirow{2}{*}{Car} &
    \multirow{2}{*}{Ped.} &
    \multirow{2}{*}{Cyc.} & 
    Latency\\
    &&&&(ms)\\
    \midrule
    1 & 92.53 & 69.15 & 89.19 & 72 \\
    2 & 92.86 & 69.50 & 93.05 & 113 \\
    3 & 92.89 & 69.70 & 93.33 & 152 \\
    4 & 92.91 & 70.04 & 94.18 & 185 \\
    5 & 93.02 & 70.00 & 92.43 & 232 \\
    
    \bottomrule
    \end{tabularx}
    \caption{Ablation studies on the KITTI \textit{val} set with AP (R40) for easy difficulty with varying sample steps.}
    \label{table:ablation_step}
\end{table}

We also conduct experiments to investigate the impacts of increasing the number of sampling steps on performance and inference time.
As a baseline model, DiffRef3D-V trained in the KITTI \textit{train} set is employed.
Results of the ablation study are summarized in Table \ref{table:ablation_step}.
Increasing the sampling steps from 1 to 5 provides an AP (R40) gain of 0.49\%, 0.85\%, and 2.24\% in \textit{car}, \textit{cyclist}, and \textit{pedestrian}, respectively.
However, AP gain introduced by increasing the number of sampling steps become insignificant compared to increase of latency after sampling steps of 3.
Therefore, we select 3 as an optimal number of sampling steps throughout the other experiments.

\subsection{Qualitative Analysis}
\begin{figure}[t]
    \centering
    \includegraphics[width=\columnwidth]{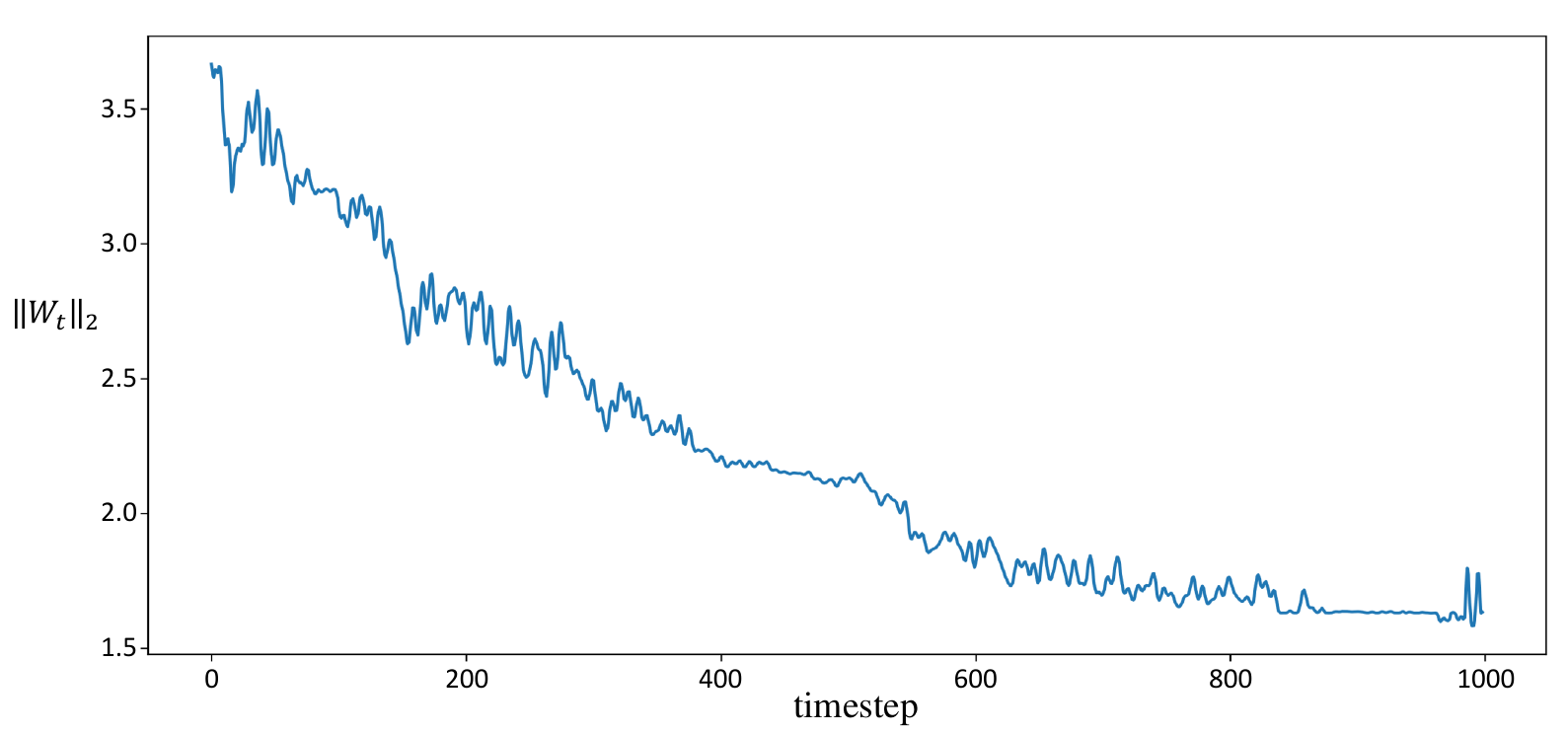}
    \caption{\textbf{Norm of scaling factor from temporal transformation}. The impact of the hypothesis feature increases as the timestep decreases after temporal transformation.}
    \label{fig4_weight_normalization}
\end{figure}

We visualize the hypothesis-driven box prediction process facilitated by the proposal refinement process of DiffRef3D. 
Figure \ref{fig3_kitti_visualization} illustrates iterative sampling steps of DiffRef3D in each timestep.
Visualization result implies the necessity of proposals as conditional inputs due to point cloud sparsity and relatively small object size compared to point cloud scenes.
Figure \ref{fig3_kitti_visualization} (a) shows hypotheses generated around proposal centers tend to prioritize regions with higher object presence, while hypotheses exhibit less realistic box forms since generated from random Gaussian noise. 
During the iterative sampling, DiffRef3D progressively refines proposals and hypotheses toward object boxes, as demonstrated in Fig. \ref{fig3_kitti_visualization} (b) and (c). 
This result highlights DiffRef3D successfully employing the diffusion process to iteratively denoise hypotheses and refine proposals.

To verify the temporal transformation block considers the effect of the noise signal, we visualize the norm of the scaling factor along every timestep.
Results are illustrated in Fig. \ref{fig4_weight_normalization}.
We affirm that the norm of the scaling factor is larger at a smaller timestep, implying that the refinement module takes more attention to the hypothesis feature at the later refinement process.
Therefore, the temporal transformation block controls the importance of hypothesis features before aggregating it with proposal features.

\section{Conclusion}
In this article, we propose a novel framework, DiffRef3D, that utilizes the diffusion process for the task of 3D object detection with point clouds.
The HAM and hypothesis generation scheme we proposed demonstrated consistent performance improvements across different models by integrating the proposal refinement of a two-stage 3D object detector into the diffusion process.
Experiments on the KITTI benchmark show that DiffRef3D can serve as a generalizable proposal refinement framework that can be applied to other two-stage 3D object detectors with simple implementation but effective performance improvement.


\bibliography{aaai24}

\begin{thebibliography}{36}
\providecommand{\natexlab}[1]{#1}

\bibitem[{Cai et~al.(2022)Cai, Pan, Yao, and Mei}]{3dcascadercnn_cai}
Cai, Q.; Pan, Y.; Yao, T.; and Mei, T. 2022.
\newblock 3D Cascade RCNN: High Quality Object Detection in Point Clouds.
\newblock \emph{IEEE Transactions on Image Processing}.

\bibitem[{Chen et~al.(2022)Chen, Sun, Song, and Luo}]{diffusiondet_chen}
Chen, S.; Sun, P.; Song, Y.; and Luo, P. 2022.
\newblock Diffusiondet: Diffusion model for object detection.
\newblock \emph{arXiv preprint arXiv:2211.09788}.

\bibitem[{Deng et~al.(2021)Deng, Shi, Li, Zhou, Zhang, and Li}]{voxelrcnn_deng}
Deng, J.; Shi, S.; Li, P.; Zhou, W.; Zhang, Y.; and Li, H. 2021.
\newblock Voxel r-cnn: Towards high performance voxel-based 3d object
  detection.
\newblock In \emph{Proceedings of the AAAI Conference on Artificial
  Intelligence}, volume~35, 1201--1209.

\bibitem[{Dhariwal and Nichol(2021)}]{adm_dhariwal}
Dhariwal, P.; and Nichol, A. 2021.
\newblock Diffusion models beat gans on image synthesis.
\newblock \emph{Advances in neural information processing systems}, 34:
  8780--8794.

\bibitem[{Geiger, Lenz, and Urtasun(2012)}]{kitti}
Geiger, A.; Lenz, P.; and Urtasun, R. 2012.
\newblock Are we ready for autonomous driving? the kitti vision benchmark
  suite.
\newblock In \emph{2012 IEEE conference on computer vision and pattern
  recognition}, 3354--3361. IEEE.

\bibitem[{Gu et~al.(2022)Gu, Chen, Xu, Lan, Meng, and Wang}]{diffusioninst_gu}
Gu, Z.; Chen, H.; Xu, Z.; Lan, J.; Meng, C.; and Wang, W. 2022.
\newblock Diffusioninst: Diffusion model for instance segmentation.
\newblock \emph{arXiv preprint arXiv:2212.02773}.

\bibitem[{He et~al.(2020)He, Zeng, Huang, Hua, and Zhang}]{sassd_he}
He, C.; Zeng, H.; Huang, J.; Hua, X.-S.; and Zhang, L. 2020.
\newblock Structure aware single-stage 3d object detection from point cloud.
\newblock In \emph{Proceedings of the IEEE/CVF conference on computer vision
  and pattern recognition}, 11873--11882.

\bibitem[{Ho, Jain, and Abbeel(2020)}]{ddpm_ho}
Ho, J.; Jain, A.; and Abbeel, P. 2020.
\newblock Denoising diffusion probabilistic models.
\newblock \emph{Advances in neural information processing systems}, 33:
  6840--6851.

\bibitem[{Hu, Kuai, and Waslander(2022)}]{pdv_hu}
Hu, J.~S.; Kuai, T.; and Waslander, S.~L. 2022.
\newblock Point density-aware voxels for lidar 3d object detection.
\newblock In \emph{Proceedings of the IEEE/CVF Conference on Computer Vision
  and Pattern Recognition}, 8469--8478.

\bibitem[{Ji et~al.(2023)Ji, Chen, Xie, Hong, Liu, Liu, Lu, Li, and
  Luo}]{ddp_ji}
Ji, Y.; Chen, Z.; Xie, E.; Hong, L.; Liu, X.; Liu, Z.; Lu, T.; Li, Z.; and Luo,
  P. 2023.
\newblock Ddp: Diffusion model for dense visual prediction.
\newblock \emph{arXiv preprint arXiv:2303.17559}.

\bibitem[{Koo et~al.(2023)Koo, Lee, Kim, Kim, Jeon, and Kim}]{pgrcnn_koo}
Koo, I.; Lee, I.; Kim, S.-H.; Kim, H.-S.; Jeon, W.-j.; and Kim, C. 2023.
\newblock PG-RCNN: Semantic Surface Point Generation for 3D Object Detection.
\newblock \emph{arXiv preprint arXiv:2307.12637}.

\bibitem[{Lai et~al.(2023)Lai, Duan, Dai, Li, Fu, Li, Qiao, and
  Wang}]{ddps_lai}
Lai, Z.; Duan, Y.; Dai, J.; Li, Z.; Fu, Y.; Li, H.; Qiao, Y.; and Wang, W.
  2023.
\newblock Denoising Diffusion Semantic Segmentation with Mask Prior Modeling.
\newblock \emph{arXiv preprint arXiv:2306.01721}.

\bibitem[{Lang et~al.(2019)Lang, Vora, Caesar, Zhou, Yang, and
  Beijbom}]{pointpillars_lang}
Lang, A.~H.; Vora, S.; Caesar, H.; Zhou, L.; Yang, J.; and Beijbom, O. 2019.
\newblock Pointpillars: Fast encoders for object detection from point clouds.
\newblock In \emph{Proceedings of the IEEE/CVF conference on computer vision
  and pattern recognition}, 12697--12705.

\bibitem[{Mao et~al.(2021)Mao, Xue, Niu, Bai, Feng, Liang, Xu, and
  Xu}]{votr_mao}
Mao, J.; Xue, Y.; Niu, M.; Bai, H.; Feng, J.; Liang, X.; Xu, H.; and Xu, C.
  2021.
\newblock Voxel transformer for 3d object detection.
\newblock In \emph{Proceedings of the IEEE/CVF International Conference on
  Computer Vision}, 3164--3173.

\bibitem[{Nichol and Dhariwal(2021)}]{improved_nichol}
Nichol, A.~Q.; and Dhariwal, P. 2021.
\newblock Improved Denoising Diffusion Probabilistic Models.

\bibitem[{Qi et~al.(2017{\natexlab{a}})Qi, Su, Mo, and Guibas}]{pointnet_qi}
Qi, C.~R.; Su, H.; Mo, K.; and Guibas, L.~J. 2017{\natexlab{a}}.
\newblock Pointnet: Deep learning on point sets for 3d classification and
  segmentation.
\newblock In \emph{Proceedings of the IEEE conference on computer vision and
  pattern recognition}, 652--660.

\bibitem[{Qi et~al.(2017{\natexlab{b}})Qi, Yi, Su, and Guibas}]{pointnet++_qi}
Qi, C.~R.; Yi, L.; Su, H.; and Guibas, L.~J. 2017{\natexlab{b}}.
\newblock Pointnet++: Deep hierarchical feature learning on point sets in a
  metric space.
\newblock \emph{Advances in neural information processing systems}, 30.

\bibitem[{Shan et~al.(2023)Shan, Liu, Zhang, Wang, Han, Wang, Ma, and
  Gao}]{d3dp_shan}
Shan, W.; Liu, Z.; Zhang, X.; Wang, Z.; Han, K.; Wang, S.; Ma, S.; and Gao, W.
  2023.
\newblock Diffusion-Based 3D Human Pose Estimation with Multi-Hypothesis
  Aggregation.
\newblock \emph{arXiv preprint arXiv:2303.11579}.

\bibitem[{Sheng et~al.(2021)Sheng, Cai, Liu, Deng, Huang, Hua, and
  Zhao}]{ct3d_sheng}
Sheng, H.; Cai, S.; Liu, Y.; Deng, B.; Huang, J.; Hua, X.-S.; and Zhao, M.-J.
  2021.
\newblock Improving 3d object detection with channel-wise transformer.
\newblock In \emph{Proceedings of the IEEE/CVF International Conference on
  Computer Vision}, 2743--2752.

\bibitem[{Shi et~al.(2020{\natexlab{a}})Shi, Guo, Jiang, Wang, Shi, Wang, and
  Li}]{pvrcnn_shi}
Shi, S.; Guo, C.; Jiang, L.; Wang, Z.; Shi, J.; Wang, X.; and Li, H.
  2020{\natexlab{a}}.
\newblock Pv-rcnn: Point-voxel feature set abstraction for 3d object detection.
\newblock In \emph{Proceedings of the IEEE/CVF Conference on Computer Vision
  and Pattern Recognition}, 10529--10538.

\bibitem[{Shi, Wang, and Li(2019)}]{pointrcnn_shi}
Shi, S.; Wang, X.; and Li, H. 2019.
\newblock Pointrcnn: 3d object proposal generation and detection from point
  cloud.
\newblock In \emph{Proceedings of the IEEE/CVF conference on computer vision
  and pattern recognition}, 770--779.

\bibitem[{Shi et~al.(2020{\natexlab{b}})Shi, Wang, Shi, Wang, and
  Li}]{parta2_shi}
Shi, S.; Wang, Z.; Shi, J.; Wang, X.; and Li, H. 2020{\natexlab{b}}.
\newblock From points to parts: 3d object detection from point cloud with
  part-aware and part-aggregation network.
\newblock \emph{IEEE transactions on pattern analysis and machine
  intelligence}, 43(8): 2647--2664.

\bibitem[{Shi and Rajkumar(2020)}]{pointgnn_shi}
Shi, W.; and Rajkumar, R. 2020.
\newblock Point-gnn: Graph neural network for 3d object detection in a point
  cloud.
\newblock In \emph{Proceedings of the IEEE/CVF conference on computer vision
  and pattern recognition}, 1711--1719.

\bibitem[{Song, Meng, and Ermon(2020)}]{ddim_song}
Song, J.; Meng, C.; and Ermon, S. 2020.
\newblock Denoising diffusion implicit models.
\newblock \emph{arXiv preprint arXiv:2010.02502}.

\bibitem[{Sun et~al.(2021)Sun, Zhang, Jiang, Kong, Xu, Zhan, Tomizuka, Li,
  Yuan, Wang et~al.}]{sparsercnn_sun}
Sun, P.; Zhang, R.; Jiang, Y.; Kong, T.; Xu, C.; Zhan, W.; Tomizuka, M.; Li,
  L.; Yuan, Z.; Wang, C.; et~al. 2021.
\newblock Sparse r-cnn: End-to-end object detection with learnable proposals.
\newblock In \emph{Proceedings of the IEEE/CVF conference on computer vision
  and pattern recognition}, 14454--14463.

\bibitem[{Vaswani et~al.(2017)Vaswani, Shazeer, Parmar, Uszkoreit, Jones,
  Gomez, Kaiser, and Polosukhin}]{transformer_vaswani}
Vaswani, A.; Shazeer, N.; Parmar, N.; Uszkoreit, J.; Jones, L.; Gomez, A.~N.;
  Kaiser, {\L}.; and Polosukhin, I. 2017.
\newblock Attention is all you need.
\newblock \emph{Advances in neural information processing systems}, 30.

\bibitem[{Wang et~al.(2023)Wang, Cao, Anwer, Xie, Khan, and
  Pang}]{dformer_wang}
Wang, H.; Cao, J.; Anwer, R.~M.; Xie, J.; Khan, F.~S.; and Pang, Y. 2023.
\newblock DFormer: Diffusion-guided Transformer for Universal Image
  Segmentation.
\newblock \emph{arXiv preprint arXiv:2306.03437}.

\bibitem[{Wu et~al.(2022)Wu, Deng, Wen, Li, and Wang}]{casa_wu}
Wu, H.; Deng, J.; Wen, C.; Li, X.; and Wang, C. 2022.
\newblock CasA: A Cascade Attention Network for 3D Object Detection from LiDAR
  point clouds.
\newblock \emph{IEEE Transactions on Geoscience and Remote Sensing}.

\bibitem[{Xu, Zhong, and Neumann(2022)}]{btc_xu}
Xu, Q.; Zhong, Y.; and Neumann, U. 2022.
\newblock Behind the curtain: Learning occluded shapes for 3d object detection.
\newblock In \emph{Proceedings of the AAAI Conference on Artificial
  Intelligence}, volume~36, 2893--2901.

\bibitem[{Xu et~al.(2021)Xu, Zhou, Wang, Qi, and Anguelov}]{spg_xu}
Xu, Q.; Zhou, Y.; Wang, W.; Qi, C.~R.; and Anguelov, D. 2021.
\newblock Spg: Unsupervised domain adaptation for 3d object detection via
  semantic point generation.
\newblock In \emph{Proceedings of the IEEE/CVF International Conference on
  Computer Vision}, 15446--15456.

\bibitem[{Yan, Mao, and Li(2018)}]{second_yan}
Yan, Y.; Mao, Y.; and Li, B. 2018.
\newblock Second: Sparsely embedded convolutional detection.
\newblock \emph{Sensors}, 18(10): 3337.

\bibitem[{Yang et~al.(2020)Yang, Sun, Liu, and Jia}]{3dssd_yang}
Yang, Z.; Sun, Y.; Liu, S.; and Jia, J. 2020.
\newblock 3dssd: Point-based 3d single stage object detector.
\newblock In \emph{Proceedings of the IEEE/CVF conference on computer vision
  and pattern recognition}, 11040--11048.

\bibitem[{Yang et~al.(2019)Yang, Sun, Liu, Shen, and Jia}]{std_yang}
Yang, Z.; Sun, Y.; Liu, S.; Shen, X.; and Jia, J. 2019.
\newblock Std: Sparse-to-dense 3d object detector for point cloud.
\newblock In \emph{Proceedings of the IEEE/CVF international conference on
  computer vision}, 1951--1960.

\bibitem[{Yin, Zhou, and Krahenbuhl(2021)}]{centerpoint_yin}
Yin, T.; Zhou, X.; and Krahenbuhl, P. 2021.
\newblock Center-Based 3D Object Detection and Tracking.
\newblock In \emph{Proceedings of the IEEE/CVF Conference on Computer Vision
  and Pattern Recognition (CVPR)}, 11784--11793.

\bibitem[{Zhou and Tuzel(2018)}]{voxelnet_zhou}
Zhou, Y.; and Tuzel, O. 2018.
\newblock Voxelnet: End-to-end learning for point cloud based 3d object
  detection.
\newblock In \emph{Proceedings of the IEEE conference on computer vision and
  pattern recognition}, 4490--4499.

\bibitem[{Zou et~al.(2023)Zou, Zhu, Ye, and Wang}]{diffbev_zou}
Zou, J.; Zhu, Z.; Ye, Y.; and Wang, X. 2023.
\newblock DiffBEV: Conditional Diffusion Model for Bird's Eye View Perception.
\newblock \emph{arXiv preprint arXiv:2303.08333}.

\end{thebibliography}

\end{document}


\section{Box and Residual Computation}
We provide details of $\ominus, \oplus$ computation between boxes and residuals.
Conventional 3D object detectors represent 3D boxes with 7-dimensional vectors $(x,y,z,w,h,l,\theta)$.
Here, $(x,y,z)$ represents center coordinates of boxes, $(w,h,l)$ represents sizes of boxes, and $\theta$ represents yaw rotation along the z-axis.
According to box encoding function \cite{second_yan}, residual $x^{gt}$ between proposal box $X^P$ and target object box $X^T$ is formulated as follows:
\begin{align}
    x^{gt} &= X^P \ominus X^T \nonumber \\
    &= (\Delta x,\Delta y,\Delta z,\Delta w,\Delta h,\Delta l,\Delta \theta), \\
    \Delta x &= \frac{x^T-x^P}{d^P}, \Delta y = \frac{y^T-y^P}{d^P}, \Delta z = \frac{z^T-z^P}{h^P}, \\
    \Delta w &= \mathrm{log}(\frac{w^T}{w^P}), \Delta h = \mathrm{log}(\frac{h^T}{h^P}), \Delta l = \mathrm{log}(\frac{l^T}{l^P}), \\
    \Delta \theta &= \theta^T - \theta^P.
\end{align}
Here, $d^P=\sqrt{(w^P)^2+(h^P)^2}$ is the diagonal of the base of the box, the superscripts $P$ and $T$ indicate the proposal and target, respectively.
Above formulation is expressed as $x^{gt}=X^T \ominus X^P$ for brevity.
In contrast, $\oplus$ represents box transformation by residual, which is inverse of $\ominus$.

\section{Residual Normalization}
DiffRef3D samples noise from Gaussian distribution $\mathcal{N}(0,I)$ to generate noisy residuals. 
However, directly applying sampled noises as box residuals would result in unlikely hypotheses since residuals in 3D object detection have a different distribution. 
From our observation of the true residual distribution, we propose to adaptively normalize the residuals with respect to the shape and size of the proposal.
Accordingly, we formulate box normalization using the notations above as follows:
\begin{align}
    x^{gt}_{norm} = (\frac{\Delta x}{d^P}, \frac{\Delta y}{d^P}, \frac{\Delta z}{l^P}, \frac{\Delta w}{d^P},\frac{\Delta h}{d^P},\frac{\Delta l}{l^P},\frac{\Delta \theta}{r^P})
\end{align}
Here, $r^P=\mathrm{max}(w^P/h^P,h^P/w^P)$ is the aspect ratio of the base of the proposal.

\section{Training Losses}
For two-stage 3D object detectors, the first stage region proposal network (RPN) and the second stage proposal refinement module are trained with independent training losses, respectively.
As the main contribution of DiffRef3D lies on the second stage, we provide the details of training losses for our proposal refinement module here.
DIffRef3D-V, PV, and T adopts the same training losses of the respective baselines.
For all three models, the classification loss $L_{cls}$ and the regression loss $L_{reg}$ is employed to supervise the confidence and bounding box prediction $\hat{c}$ and $\hat{x}$, respectively.

For regression loss $L_{reg}$, the smooth-L1 loss is computed in two terms: residuals and box corner points. 
It is formulated as follows:
\begin{multline}
    L_{reg} = \mathbb{1}(\mathrm{IoU}_i \geq \theta_{reg}) \frac{1}{N_P}\sum^{N_P}_{i=1}[\mathrm{smoothL1}(x^{gt}_i, \hat{x}^{(t)}_i) \\ + \sum^8_{j=1}\mathrm{smoothL1}(b^T_{i,j}, b^P_{i,j})].
\end{multline}
$\mathrm{IoU}_i$ represents $\mathrm{IoU}$ between $i$-th proposal and corresponding target box and $\mathbb{1}(\mathrm{IoU}_i \geq \theta_{reg})$ indicates that only region proposals with $\mathrm{IoU}$ over threshold $\theta_{reg}$ contributes to the regression loss, $N_P$ is number of proposals. 
For corner loss, $b_{i,j}$ indicates $j$-th corner point coordinates, and superscripts $T$ and $P$ indicate the target object box and proposal box, respectively.

In case of classification loss $L_{cls}$, the target is assigned by an $\mathrm{IoU}$ related value.
Specifically, the target for $i$-th proposal is assigned as follows:
\begin{equation}
c^{gt}_i = \begin{cases}
    1 & \mathrm{IoU}_i \geq \theta_H, \\
    \frac{\mathrm{IoU}_i-\theta_L}{\theta_H-\theta_L} & \theta_L \leq \mathrm{IoU}_i<\theta_H, \\ 
    0 & \mathrm{IoU}_i < \theta_L,
\end{cases}
\end{equation}
and the binary cross-entropy loss is computed according to assigned target as follows:
\begin{equation}
    L_{cls} = \frac{1}{N_P}\sum^{N_P}_{i=1}L_{BCE}(c^{gt}_i, \hat{c}_i).
\end{equation}

DiffRef3D-PV employs additional keypoint segmentation loss \cite{pvrcnn_shi} to the above losses, which is implemented with a Focal loss \cite{focal_yun} supervised with the foreground label for each keypoint.
Formally, the segmentation loss $L_{seg}$ is formulated as:
\begin{equation}
    L_{seg} = -\frac{1}{N_{kp}}\sum^{N_{kp}}_{k=1}L_{Focal}(s_{k}^{gt}, \hat{s}_k)
\end{equation}
where $N_{kp}$ is the number of keypoints, $\hat{s}_k$ denotes the foreground score prediction.
The foreground label $s^{gt}_{k}$ is assigned as 1 if the keypoint lies inside an object bounding box, otherwise 0.

\begin{figure*}
    \centering
    \includegraphics[width=\textwidth]{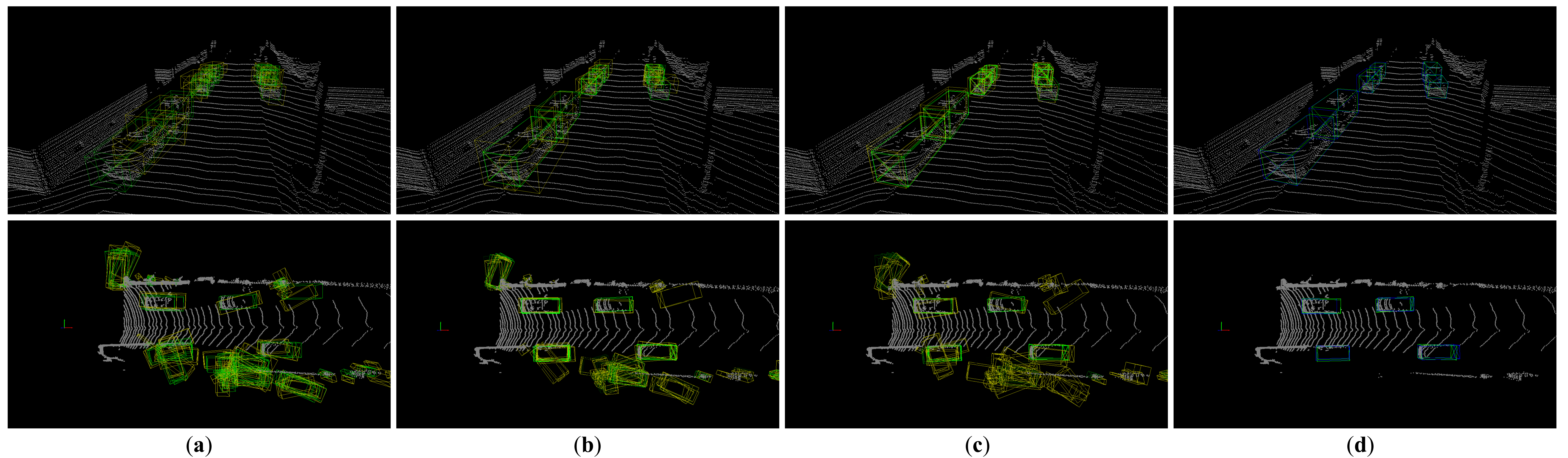}
    \caption{\textbf{More qualitative results.} Visualizations of iterative sampling steps of DiffRef3D on various scenes. Each image shows hypotheses and predictions at (a) t=1000, (b) t=500, (c) t=200, and (d) the final prediction. Hypotheses are colored in yellow, predictions are colored in green, and ground truths are colored in blue.}
    \label{figs1_qualitative_results}
\end{figure*}

\section{Additional Experiments}

\begin{figure}
    \centering
    \includegraphics[width=\columnwidth]{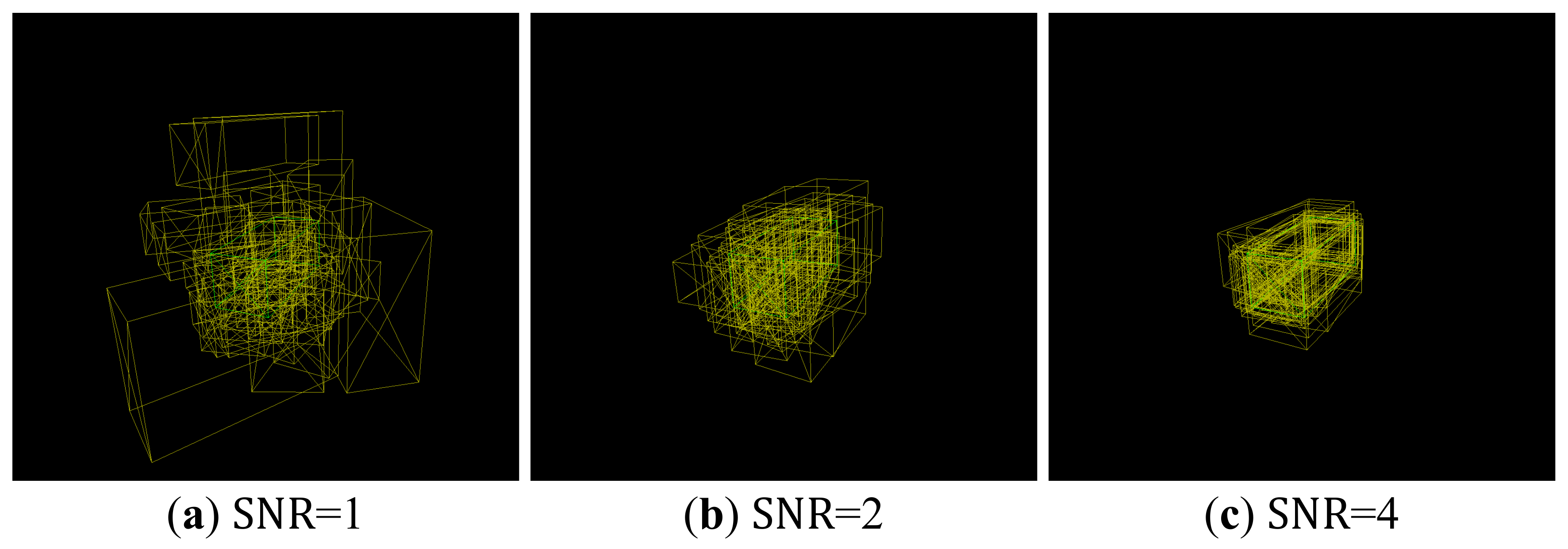}
    \caption{\textbf{Hypotheses generation on different SNRs.} We visualize how hypotheses are generated based on reference boxes according to various SNRs at timestep $t=1000$. Hypotheses are colored in yellow, reference boxes are colored in green.}
    \label{figs2_snr}
\end{figure}

In this section, we present more qualitative results and ablation studies on the KITTI benchmark \cite{kitti}. 
Figure \ref{figs1_qualitative_results} shows visualization results on the inference phase in various scenes.

\paragraph{Signal-to-Noise Ratio}
\begin{table}[h]
    \centering
    \caption{Ablation studies with different signal-to-noise ratio on the KITTI \textit{val} set with AP (R40) for moderate difficulty.}
    \label{table:ablation_snr}
    \begin{tabularx}{0.9\columnwidth}{Y|YYYYYY}
\toprule
\multirow{2}{*}{SNR} & \multicolumn{2}{c}{Car} & \multicolumn{2}{c}{Pedestrian} & \multicolumn{2}{c}{Cyclist} \\
 & {\small step1} & {\small step3} &{\small step1} & {\small step3} & {\small step1} & {\small step3} \\
\midrule
1 & 77.80 & 77.89 & 61.36 & 62.83 & 68.63 & 69.44 \\
2 & \textbf{84.98} & \textbf{84.77} & 60.45 & 63.07 & \textbf{71.80} & \textbf{74.20} \\
4 & 83.47 & 83.47 & \textbf{62.47} & \textbf{63.44} & 71.57 & 72.39 \\
\bottomrule
    \end{tabularx}
\end{table}
We further study the influence of signal-to-noise ratio (SNR) in Table \ref{table:ablation_snr}.
Results demonstrate that the SNR of 2 achieves optimal AP performance on the KITTI \textit{val} set.
We explain that performance degradation at SNR of 1 is due to impractical shapes of hypotheses that fall short on detection.
Figure \ref{figs2_snr} illustrates how hypotheses are generated based on proposals according to each SNR.
Hypotheses generated by SNR of 1 could appear far from proposal which leverages irrelevant features, and SNR of 4 generates hypotheses condensed on proposals which lack to exploit additional local features.
Note that, Table \ref{table:ablation_snr} highlights AP improvements at larger sampling steps in general regardless of SNR, which implies the effectiveness of iterative sampling steps by the diffusion process on 3D object detection.

\paragraph{Attention Mechanism}
\begin{table}[ht]
\centering
\caption{Ablation studies with different implementation of attention mechanism on the KITTI \textit{val} set with AP (R40) for moderate difficulty. Method "MLP" stands for employing a multi-layer perceptron rather than the attention block, "C.A." stands for the cross-attention between proposal and hypothesis feature, and "S.A." stands for the self-attention on hypothesis feature, respectively. Involving $x_t$ indicates embedding positional information from noisy residuals into features or queries.}
\label{table:ablation_attn}
\begin{tabularx}{0.9\columnwidth}{cc|YYY}
\toprule
\multirow{2}{*}{Method} & Involve & \multicolumn{3}{c}{Mod. AP(R40), step3} \\
 & $x_t$ & Car & Ped. & Cyc. \\
\midrule
\multirow{2}{*}{MLP} &  & 82.37 & 62.23 & 72.28 \\
 & \ding{51} &  82.48 & 61.49 & 73.68\\
\midrule
\multirow{2}{*}{C.A.} & & 82.56 & 62.66 & 72.52 \\
 & \ding{51} & 82.82 & 63.81 & 72.79 \\
\midrule
\multirow{2}{*}{S.A.} &  & 83.44 & \textbf{64.62} & 72.92 \\
 & \ding{51} & \textbf{84.77} & 63.07 & \textbf{74.20} \\
\bottomrule
\end{tabularx}
\end{table}

We conduct experiments to further verify design choice of attention mechanism in hypothesis attention module (HAM).
Table \ref{table:ablation_attn} demonstrates results of various implementations, on the KITTI \textit{val} set with sampling steps of 3.
Only with self-attention already outperforms other methods involving $x_t$ on every class, and AP gains derived from involving $x_t$ is higher at self-attention compared to other methods.
Therefore, we designed our HAM to include a self-attention block involving $x_t$ as positional embedding on query.

\paragraph{Temporal Transformation Block}
As shown in Fig. \ref{figs1_qualitative_results}, hypotheses generated at smaller timestep are located closer to the predicted object bounding box.
Figure 4 in the main manuscript also highlights that our temporal transformation block regularizes the contribution of the hypothesis feature with respect to timestep.
Table \ref{table:ablation_tt} quantitatively presents the effect of the temporal transformation block, verifying its necessity.

\begin{table}[h]
    \caption{Ablation studies with temporal transformation block on the KITTI \textit{val} set with AP (R40) for moderate difficulty.}
    \label{table:ablation_tt}
    \centering
    \begin{tabularx}{0.9\columnwidth}{Y|YYYYYY}
\toprule
\multirow{2}{*}{TT} & \multicolumn{2}{c}{Car} & \multicolumn{2}{c}{Pedestrian} & \multicolumn{2}{c}{Cyclist} \\
 & {\small step1} & {\small step3} &{\small step1} & {\small step3} & {\small step1} & {\small step3} \\
\midrule
 & 77.80 & 77.89 & \textbf{61.36} & 62.83 & 68.63 & 69.44 \\
\ding{51} & \textbf{84.98} & \textbf{84.77} & 60.45 & \textbf{63.07} & \textbf{71.80} & \textbf{74.20} \\
\bottomrule
    \end{tabularx}
\end{table}

\begin{table}[ht]
\centering
\caption{Ablation studies with box ensemble methods on the KITTI \textit{val} set with AP (R40) for moderate difficulty \textit{pedestrian} class. Method "None" stands for no ensemble applied during the iterative sampling steps, "NMS" stands for the non-maximum suppression applied on ensemble of box predictions then applied averaging, and "Mean" stands for the averaging ensemble of box predictions, respectively.}
\label{table:ablation_ensemble}
\begin{tabularx}{0.9\columnwidth}{Y|YYY}
\toprule
\multirow{2}{*}{Ensemble} & \multicolumn{3}{c}{Sampling Steps} \\
& 1 & 2 & 3 \\
\midrule
None & 60.45 & 62.70 & 61.13 \\
NMS & 60.45 & 60.58 & 60.56 \\
Mean & 60.45 & 62.77 & 63.07 \\
\bottomrule
\end{tabularx}
\end{table}

\paragraph{Box Ensemble}
We further compare the box ensemble methods during the inference phase.
Table \ref{table:ablation_ensemble} summarizes the comparison results.
Notably, non-maximum suppression (NMS) results lower AP (R40) compared to no ensemble applied.
Box ensemble via averaging outperforms other methods along each sampling step, thus we adopted the method to boost detection performance. 

\bibliography{aaai24}